\documentclass[11pt]{article}

\usepackage[final]{acl}

\usepackage{amsfonts} 
\usepackage{times}
\usepackage{latexsym}

\usepackage[T1]{fontenc}
\usepackage[utf8]{inputenc}

\usepackage{microtype}

\usepackage{inconsolata}

\usepackage{graphicx}
\usepackage{amsmath}
\usepackage{multirow}
\usepackage{placeins}
\usepackage{listings}
\title{Exploring Health Misinformation Detection with Multi-Agent Debate}

\author{
  Chih-Han Chen\thanks{Work done while interning at HTC DeepQ.} \\
  National Taiwan University \\
  \texttt{chen.eric0208@gmail.com}
  \And
  Chen-Han Tsai \\
  HTC DeepQ \\
  \texttt{maxwell\_tsai@htc.com} 
  \And
  Yu-Shao Peng \\
  HTC DeepQ \\
  \texttt{ys\_peng@htc.com}
}

\begin{document}
\maketitle
\begin{abstract}
% Fact-checking health-related claims is increasingly crucial as large amounts of misinformation circulate on the web. Effective verification requires retrieving high-quality articles and providing sound reasoning. In this paper, we propose a two-stage approach: Agreement Score Prediction followed by Multi-Agent Debate. In the first stage, we leverage large language models (LLMs) to assess retrieved articles individually and compute an aggregated agreement score. When the agreement score does not meet a preset threshold, the process advances to a second stage, where agents engage in a structured debate to reach a supported decision with explicit reasoning. Experimental results demonstrate that the two-stage approach outperforms baseline methods.

%Max-Claude Version
Fact-checking health-related claims has become increasingly critical as misinformation proliferates online. Effective verification requires both the retrieval of high-quality evidence and rigorous reasoning processes. In this paper, we propose a two-stage framework for health misinformation detection: Agreement Score Prediction followed by Multi-Agent Debate. In the first stage, we employ large language models (LLMs) to independently evaluate retrieved articles and compute an aggregated agreement score that reflects the overall evidence stance. When this score indicates insufficient consensus—falling below a predefined threshold—the system proceeds to a second stage. Multiple agents engage in structured debate to synthesize conflicting evidence and generate well-reasoned verdicts with explicit justifications. Experimental results demonstrate that our two-stage approach achieves superior performance compared to baseline methods, highlighting the value of combining automated scoring with collaborative reasoning for complex verification tasks.

\end{abstract}

\begin{figure*}[t]
    \centering
    \includegraphics[width=1\textwidth]{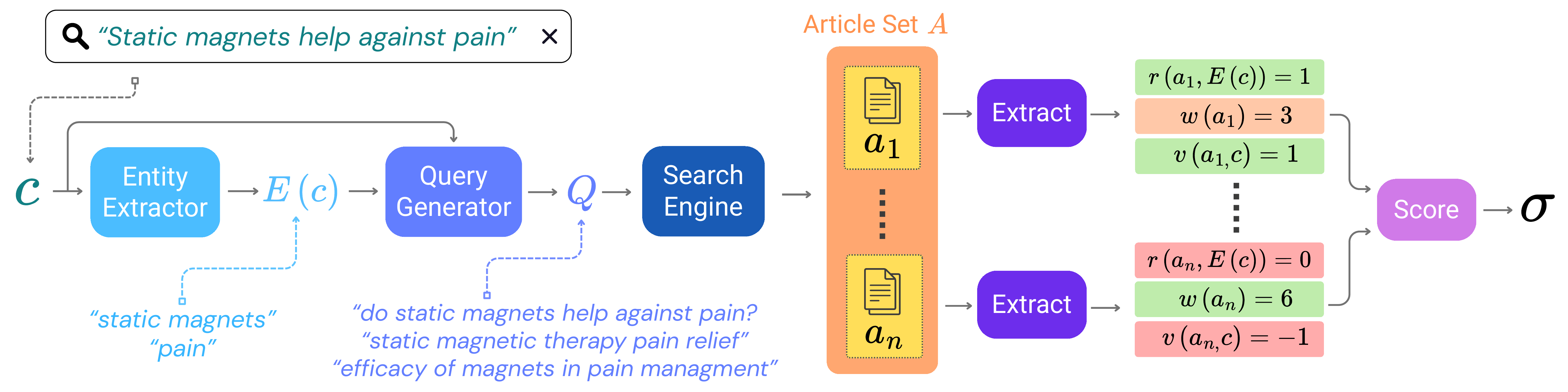}
    \caption{\textbf{Agreement Score Prediction (Stage 1).} For a given claim $c$, entities $E(c)$ are extracted and passed alongside $c$ to a query generator to generate search queries $Q$. Articles relating to $Q$ are collected into an article set $A$. We extract topic relevance $r$, article weights $w$, and article verdict $v$ for each article $a \in A$. The results are aggregated, resulting in the final agreement score $\sigma$.}
    \label{fig:stage1}
\end{figure*}

\section{Introduction \& Related Work}

The proliferation of health-related content on digital platforms poses significant challenges to ensuring accurate medical information reaches the public. Verifying health claims is critical for safeguarding public well-being, as false or misleading information can cause substantial harm to individual and population health. Despite the vast volume of health content available online, only a small fraction is supported by robust scientific evidence, underscoring the urgent need for automated verification systems.

In open-domain fact-checking, traditional methods predominantly rely on BERT-based architectures \cite{devlin-etal-2019-bert}. Pipeline-based systems employ BERT models to retrieve relevant evidence sentences, followed by a classification module to predict claim veracity. Joint systems perform evidence retrieval and veracity prediction simultaneously within a unified model. While conceptually straightforward, these approaches require predefined knowledge databases and necessitate training encoder-based models from scratch \cite{vladika-etal-2024-healthfc}, limiting their flexibility and scalability.

The emergence of large language models (LLMs) has introduced new paradigms. \citet{tian2024web} deploy web retrieval agents to gather evidence dynamically, enabling LLMs to assess sufficiency and render verdicts. \citet{singal-etal-2024-evidence} integrate retrieval-augmented generation (RAG) with in-context learning (ICL) for veracity prediction. \citet{vladika-etal-2025-step} propose multi-turn LLM interactions that iteratively generate questions, retrieve evidence, and reason about claim validity. However, these approaches typically lack explicit evidence filtering mechanisms, relying directly on outputs from web search tools or dense retrieval models.

Recent work has explored multi-agent frameworks for fact-checking. \citet{hong2025emulatemultiagentframeworkdetermining} leverage multiple agents to evaluate evidence quality and determine veracity, with provisions for re-gathering evidence when necessary. \citet{hu-etal-2025-removal}, \citet{liang-etal-2024-encouraging}, and \cite{Liu2025TheTB} adopt Multi-Agent Debate (MAD) frameworks to enhance reasoning robustness and mitigate degenerate reasoning patterns.

Building upon these advances, we propose a two-stage multi-agent debate framework for health misinformation detection. Our approach first employs LLMs to retrieve and evaluate high-quality articles, computing an aggregated agreement score. When evidence exhibits significant disagreement—indicated by a score below a predefined threshold—the system initiates a structured multi-agent debate. Through iterative argumentation, agents collaboratively analyze conflicting evidence to produce well-justified verdicts grounded in explicit reasoning.

\section{Methodology}

In this section, we detail the implementation of our proposed two-stage health misinformation detection algorithm. The first stage takes a claim as input and retrieves a set of articles relating to the claim. Each article is classified as to whether it \emph{Supports} or \emph{Refutes} the claim, and the predictions are aggregated. When the agreement among predictions is high, the veracity of the claim is determined by majority vote. In the case of low agreement, we initiate the second stage multi-turn debate. Two opposing agents are provided with supporting evidence collected during the first stage, and a judge agent supervises the debate process until the claim's veracity can be determined. The details of each stage are presented in the following.

% ,  retrieves a set of evidence articles, classifies each article as `supporting' or `refuting' the claim

% aggregates LLM-scored support/refute signals into a weighted-majority verdict; when agreement is low, we invoke a multi-agent debate to produce a final, explicitly reasoned decision.

% \textcolor{red}{(Fixed) What does Multi-Agent Voting and Debate each do? A one liner describing the high-level purpose of each of these stages suffice.}

\subsection{Agreement Score Prediction}
% The goal of the first stage is to predict an \textit{agreement score} $\sigma \in [-1,1]$ for a given input claim $c$. 
% In the first stage, query expansion is performed on a given claim $c$ to obtain query set $Q$. Query set $Q$ is passed to a search engine, which outputs a set of articles $A$.
% Each article $a \in A$ is assessed following a series of criteria (elaborated below), and the results are aggregated to produce the final agreement score $\sigma \in [-1, 1]$. The details are provided in the following.
% from a set of evidence articles $A$. Each article $a \in A$ is evaluated as to whether it \emph{Supports} or \emph{Refutes} the claim $c$. The evaluations are aggregated to produce a final agreement score $\sigma$. Claims with high agreement are directly assigned the majority vote evaluated using $A$. Claims with a low agreement score are passed to the second Debate stage for further evaluation. We discuss the implementation of our evidence retrieval and majority agreement mechanism in the following.

% using 
% retrieve up to $N$ high-quality evidence articles from the web and determine, for each, whether the evidence supports or refutes the claim. We compute a confidence score for the retrieved evidence. If the score is sufficiently strong to support or refute the claim, the system directly outputs the prediction. Otherwise, the process proceeds to the second stage.

\begin{figure*}[t]
    \centering
    \includegraphics[width=1\textwidth]{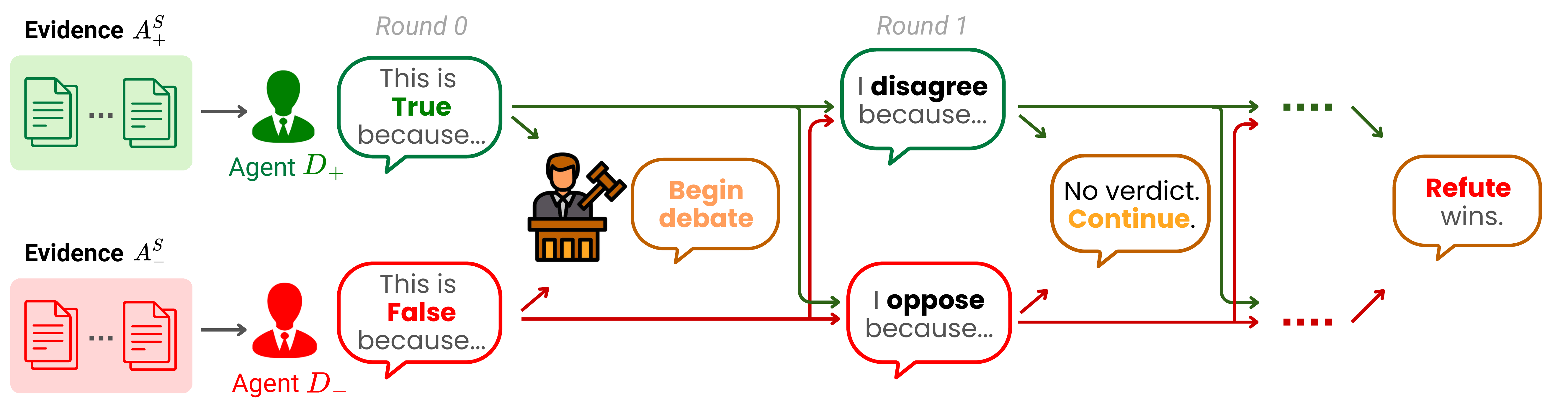}
    \caption{\textbf{Multi-Agent Debate (Stage 2).} Articles from the first Agreement Score Prediction stage are organized into supporting and refuting evidence sets $A^S_+$ and $A^S_-$, which are provided to agents $D_+$ and $D_-$, respectively. Each agent begins with an opening statement based on their evidence set, after which the judge initiates the debate. In each round, agents review their opponent's argument before providing a counterargument. After each round, the judge determines whether sufficient information exists to reach a verdict. If not, the debate continues for another round. The process concludes when the judge reaches a final verdict.}
    \label{fig:stage2}
\end{figure*}

% \subsubsection{Article Retrieval}

Figure~\ref{fig:stage1} illustrates the first stage framework of our approach. For a given claim $c$, we first extract a set of entities $E(c)$ from $c$ using an LLM. The entities are keywords or phrases from $c$ that the claim is focused on. The claim $c$ and entities $E(c)$ are then provided to an LLM to generate a set of queries $Q$. 
% The first query $q_0$ is claim $c$ rephrased as a natural question. The remaining queries $q_{>0}$ are topic decompositions of $c$ with a focus on the entities. 
Each query $q \in Q$ is sent to a search engine for article retrieval. The article sets retrieved from each query are de-duplicated and merged to form the article set $A$.

% To retrieve the most relevant articles, we first apply an entity extractor $E$ to identify the two entities $E_1$ and $E_2$ in a claim $c$. 
% Since health claims in our dataset consistently follow this format, extracting these entities allows us to verify whether retrieved articles provide information that enriches or clarifies the claim. 
% This extraction step is performed using an LLM.

% Given a claim $c$ with entities $e_1$ and $e_2$, a query generator $G$ produces five search queries: one phrased as a natural question $q$, and four derived through topic decomposition. 
% This generation step is performed using an LLM. 
% The resulting five queries are then issued to a web search engine, yielding a candidate article set $A$.

% \subsubsection{Article Assessment}

Given the obtained queries $Q$, entities $E(c)$, and article set $A$, we prompt an LLM to extract the following information from each article $a \in A$. Specifically, we look for:
% First, we extract the \textbf{relevance} $r(a, E) \in \{0,1\}$ of each article using the extracted entities $E$. Specifically, $r(a,E)=1$ if all entities in $E$ have been mentioned in article $a$. 

\begin{enumerate}
    \item \textbf{Topic Relevance:} Check whether the article $a$ contains content relevant for all entities in $E(c)$.
    % all entities in $E$ or their semantic meanings are mentioned in article $a$. 
    We define this relevance as $r(a, E(c)) \in \{0,1\}$, where $r(a, E(c))=1$ if the article contains content relevant for all entities in $E(c)$ and $r(a, E(c))=0$ otherwise.
        
    \item \textbf{Attribute Assessment:} Evaluate whether article $a$ contains the following attributes: \textit{Problem Statement}, \textit{Experimental Setup}, \textit{Findings}, \textit{Statistical Significance}, \textit{Limitations}, and \textit{Results}. These $6$ attributes reflect the structure of modern scientific publications. Specifically, an article that coverts the $6$ attributes are often more thorough in its claims. We define the article weight as:
    \[
    w(a) = \sum_{\alpha \in \text{Attributes}} \mathbf{1}[\alpha \in a] \;\in \{0,1,\dots,6\}
    \]
    where $\mathbf{1}[\cdot]$ is the indicator function for whether attribute $\alpha$ is in article $a$.
    
    \item \textbf{Article Verdict:} Determine whether the contents of the article $a$ \emph{support} or \emph{refute} the claim $c$. We denote $v(a, c) \in \{-1,1\}$ where $v(a,c)=1$ indicates \emph{support} and $v(a,c)=-1$ indicates \emph{refute}.
\end{enumerate}

We then compute the \textit{agreement score} $\sigma(c,A) \in [-1, 1]$ for claim $c$ and article set $A$ as:
\[
\sigma(c, A) = \frac{1}{Z} \sum_{a \in A} r(a, E(c)) \cdot w(a) \cdot v(a, c),
\]
where
\[
Z = \sum_{a \in A} r(a, E(c)) \cdot w(a)
\]

is the normalizing constant. We consider the case where $Z \neq 0$ by assuming quality relevant articles to be available within the search engine results.

We introduce a threshold $\tau > 0$ to quantify the \textit{level of agreement} among the retrieved articles. If $\lvert \sigma \rvert \geq \tau$, this indicates that most articles consistently \emph{support} or \emph{refute} the claim. When such high level of agreement exists, the first stage directly outputs \textit{support} for $\sigma \ge \tau$ and \textit{refute} for $\sigma \le -\tau$. 

Conversely, an agreement score $\lvert \sigma \rvert < \tau$ indicates a significant level of disagreement among the articles. In this case, we pass the results to the second stage for debate.

% The article set $A$, rephrased query $q_0$, relevance scores $\{r\}$, attribute scores.

% and the article verdicts $\{v(a,c)\} \;\forall a\in A$ are  sent to the second stage for debate. 

% $\operatorname{sign}(x)$

% We define $\lvert \sigma \rvert$ as the strength of agreement and introduce a threshold $\tau$ to quantify the degree of agreement among the retrieved evidence. If $\lvert \sigma \rvert \geq \tau$, the score $\sigma$ is sufficiently far from $0$, indicating that most articles consistently support or refute the claim. We regard this as a case of high agreement, and the system directly outputs the prediction $\hat{y}$.  Conversely, if $\lvert \sigma \rvert < \tau$, the score remains close to $0$, reflecting uncertainty or disagreement among the articles. In such low-agreement cases, the process proceeds to the second stage for further resolution.

% \begin{equation}
% \hat{y} =
% \begin{cases}
% \text{support}, & \text{if } \lvert \sigma \rvert \geq \tau \ \text{and } \sigma > 0, \\
% \text{refute},  & \text{if } \lvert \sigma \rvert \geq \tau \ \text{and } \sigma < 0, \\
% \text{debate},  & \text{otherwise}.
% \end{cases}
% \end{equation}

\subsection{Multi-Agent Debate}

Figure~\ref{fig:stage2} illustrates the second stage framework of our approach. We employ a multi-agent debate framework based on the work by \citet{liang-etal-2024-encouraging}.
% to reason about the final prediction for the debate topic, i.e., the question $q$.
The debate framework involves three agents: the \textbf{Support Agent} $D_+$, \textbf{Refute Agent} $D_-$, and \textbf{Judge Agent} $J$. Evidence is first prepared using the results from the first stage before initiating the debate.

\paragraph{Evidence Preparation:}  
Given the article set $A$, we select two disjoint subsets $A_+$ and $A_-$ from $A$ such that:
\[
\begin{aligned}
A_+ &= \{a \in A \mid v(a,c)=+1,\; r(a,E(c))=1 \}, \\
A_- &= \{a \in A \mid v(a,c)=-1,\; r(a,E(c))=1 \}.
\end{aligned}
\]
Articles in $A_+$ and $A_-$ are ranked in descending order using $w(a)$, and we limit each set to contain an equal number of articles. For each article in the remaining sets, we prompt an LLM using the claim $c$ to extract passages from the original text that \emph{supports} or \emph{refutes} claim $c$ along with its reason.
We concatenate the LLM responses from all articles in sets $A_+$ and $A_-$ into $A^{s}_{\text{+}}$ and $A^{s}_{\text{-}}$. We denote $A^{s}_{\text{+}}$ and $A^{s}_{\text{-}}$ as the \textit{supporting} and \textit{refuting evidence} throughout the debate process.

% From the first stage, we obtain the sets of articles that are relevant to the claim $c$ and either support or refute it, denoted as $A_{\text{+}}$ and $A_{\text{-}}$, respectively.  
% We rerank each set by the attribute weight $w(a)$ and then assign an equal number of articles to $D_{\text{+}}$ and $D_{\text{-}}$.  
% For each article, we perform evidence extraction, which yields one or more snippets consisting of: (i) the section where the evidence appears, (ii) the original context, and (iii) reasoning on how the context addresses $q$.  
% The resulting supporting and refuting evidence sets are denoted as $P_{\text{+}}$ and $P_{\text{-}}$.  

\paragraph{Opening Statement:}
The support agent $D_{\text{+}}$ and refute agent $D_{\text{-}}$ begins with an opening statement by presenting the evidence in $A^S_+$ and $A^S_-$. We denote the outputs of the support and refute agents as
\[
S^{(0)}_{\text{+}} = D_{\text{+}}(A^{s}_{\text{+}}), \quad
S^{(0)}_{\text{-}} = D_{\text{-}}(A^{s}_{\text{-}}).
\]

Each agent also maintains a conversation history $H$. Following the opening statement, we initialize each agent's history as
\[
H^{(0)}_+ = \{ S^{(0)}_{\text{+}} \}, \quad
H^{(0)}_- = \{ S^{(0)}_{\text{-}} \}.
\]
The judge agent's history is initialized using the opening statements given by the two debate agents
\[
H^{(0)}_\text{J} = \{ S^{(0)}_{\text{+}} , S^{(0)}_{\text{-}} \}. \quad
\]
Next, the judge initiates the debate process, and we proceed to the first round of debate.

\paragraph{Debate Process:}
In every debate round, each agent responds to the opposing agent’s statement $S^{(i-1)}$ using its past conversation history $H^{(i-1)}$.
The outputs of the support and refute agent from the $i$-th round are given as
% $H^{(r-1)}_{\text{+}}$, $H^{(r-1)}_{\text{-}}$, and $H^{(r-1)}_{\text{j}}$ 
% denote the cumulative histories of $D_{\text{+}}$ and $D_{\text{-}}$ and $J$ , respectively, up to round $r-1$. We define:
\[
\begin{aligned}
S^{(i)}_{\text{+}} &= D_{\text{+}}\big(S^{(i-1)}_{\text{-}}, H^{(i-1)}_{\text{+}}\big), \\
S^{(i)}_{\text{-}} &= D_{\text{-}}\big(S^{(i-1)}_{\text{+}}, H^{(i-1)}_{\text{-}}\big).
\end{aligned}
\]

The debate agent's histories are updated by concatenating the opposing agent's response along with the current response
\[
\begin{aligned}
H^{(i)}_{\text{+}} &= H_+^{(i-1)} \oplus S_-^{(i-1)} \oplus S_+^{(i)} , \\
H^{(i)}_{\text{-}} &= H_-^{(i-1)} \oplus S_+^{(i-1)} \oplus S_-^{(i)}.
\end{aligned}
\]

The judge agent $J$ takes the response from both agents along with its own history $H_\text{J}^{(i-1)}$, and decides whether sufficient information exists to reach a verdict. Specifically, 
\[
\begin{split}
\theta^{(i)} &= J\Bigl( S^{(i)}_{\text{+}},\; S^{(i)}_{\text{-}},\; H^{(i-1)}_{\text{J}} \Bigr)\,
% j^{(r)} &\in \{D_{\text{+}}, D_{\text{-}}, \text{undecided}\}.
\end{split}
\]
where $\theta^{(i)}\in\{\texttt{support},\texttt{refute},\texttt{continue}\}$.
If the judge agent believes an argument is compelling enough, the verdict $\theta^{(i)} \in  \{\texttt{support},\texttt{refute} \}$ is returned.
If neither argument is sufficiently convincing, the judge agent outputs $\theta^{(i)} = \texttt{continue}$, and the debate continues for another round. 

The judge's history is also updated by appending the debate agent responses
\[
\begin{aligned}
H^{(i)}_{\text{J}} &= H_\text{J}^{(i-1)} \oplus S_+^{(i)} \oplus S_-^{(i)}. \\
\end{aligned}
\]

To prevent indefinitely long debates, we limit the process to a maximum of $M$ rounds, after which the judge must reach a verdict $\theta^{(M)} \in \{\texttt{support}, \texttt{refute}\}$ based on the debate history.
% \paragraph{Stopping Rule:}
% The final outcome $\hat{y}$ is defined as
% \begin{equation} \label{eq:stopping}
% \begin{split}
% \hat{y} = 
% \begin{cases}
% j^{(r)}, & \text{if } j^{(r)} \in \{ S^{(r)}_{\text{+}}, S^{(r)}_{\text{-}} \},\; r \leq 4, \\
% J_{\text{force}}\big(H^{(r-1)}_{\text{j}}\big), & \text{if } r = 5.
% \end{cases}
% \end{split}
% \end{equation}

% That is, if the judge reaches a decision within the first four rounds, the debate 
% terminates immediately. Otherwise, in the fifth (final) round, the judge is 
% compelled to select one side and return the final answer.

\begin{table*}[!t]
\centering
\begin{tabular}{l|ccc|ccc|ccc}
\hline
\multirow{2}{*}{\textbf{Method}} 
& \multicolumn{3}{c|}{\textbf{SciFact}} 
& \multicolumn{3}{c|}{\textbf{TREC-Health}} 
& \multicolumn{3}{c}{\textbf{HealthFC}} \\
& \textbf{P} & \textbf{R} & \textbf{F1} 
& \textbf{P} & \textbf{R} & \textbf{F1} 
& \textbf{P} & \textbf{R} & \textbf{F1} \\
\hline
\textsc{WebAgent} \cite{tian2024web} & 80.1 & 83.2 & 80.6 & 76.2 & 75.6 & 75.7 & 78.0 & 78.3 & 78.1 \\
\textsc{StepbyStep} \cite{vladika-etal-2025-step} & \textbf{86.1} & \textbf{89.5} & \textbf{87.8} & 69.9 & \textbf{95.1} & 80.6 & 72.6 & \textbf{91.6} & 81.0 \\
\textsc{Ours (1st stage only)} & 84.9 & 86.1 & 85.5 & \textbf{83.8} & 78.2 & 78.3 & 76.9 & 73.4 & 74.3 \\
\textsc{Ours (1st stage + 2nd stage)} & 82.4 & 85.3 & 83.1 & 81.3 & 81.5 & \textbf{81.4} & \textbf{82.1} & 82.7 & \textbf{82.4} \\
\hline
\end{tabular}
\caption{Performance comparison across three datasets (SciFact, TREC-Health, and HealthFC) using macro precision (P), recall (R), and F1 score. Best results are in \textbf{bold}.}
\label{tab:results}
\end{table*}

\section{Experiments and Setup}

\subsection{Datasets}
We consider the following health-related datasets for our experiments. 

\paragraph{SciFact} \cite{wadden-etal-2020-fact} contains expert-written biomedical claims derived from medical paper abstracts. We use the development subset, consisting of $188$ claims: $124$ supported and $64$ refuted.

\paragraph{TREC-Health} \cite{10.1007/978-3-031-28238-6_48} is constructed from the TREC 2019 Decision Track \cite{Abualsaud2020OverviewOT} and the TREC 2021 Health Misinformation Track \cite{Clarke2021OverviewOT}, both of which target challenges in search engine results related to health misinformation. The dataset includes $113$ consumer health questions, of which $61$ are supported and $52$ are refuted.

\paragraph{HealthFC} \cite{vladika-etal-2024-healthfc} consists of everyday health-related claims spanning diverse topics. We use a subset of $327$ claims: $202$ supported and $125$ refuted.

\subsection{Metrics}

We report macro-precision, macro-recall, and macro-F1 as evaluation metrics. These are standard in fact-checking tasks, as they provide a balanced analysis of prediction performance across labels.

\subsection{Baseline Algorithms}

We consider \textsc{WebAgent} \cite{tian2024web} and \textsc{StepbyStep} \cite{vladika-etal-2025-step} as benchmark algorithms. Among them, \textsc{StepbyStep} represents the current state-of-the-art in health-related fact-checking. For fairness, all methods, including ours, use the Brave search engine \cite{brave_api} and GPT-4o \cite{openai_gpt4o} as the underlying LLM. Each algorithm is executed three times, and we report the best performance.  

For our framework, we set the parameters as follows: entity set size $|E(c)| = 2$, query set size $|Q| = 5$, article set size $|A| = 10$, agreement threshold $\tau = 0.7$, and debate round limit $M = 5$.  

\subsection{Comparison Results \& Analysis}

The experimental results are shown in Table~\ref{tab:results}. Our first-stage-only method achieves better performance comparable to \textsc{WebAgent}, although \textsc{StepbyStep} remains challenging to surpass. 

When the second-stage debate mechanism is incorporated, our approach yields substantial improvements over the first-stage-only variant: F1 scores increase by $+3.1$ on TREC-Health and $+8.1$ on HealthFC. This demonstrates that, in cases of uncertain agreement among retrieved articles, the debate mechanism enables more effective reasoning and leads to stronger overall performance. 

Compared to \textsc{StepbyStep}, our two-stage pipeline achieves higher F1 performance by $+0.8$ on TREC-Health and $+1.4$ on HealthFC. Notably, our method maintains a balance between precision and recall, whereas \textsc{StepbyStep} tends to favor high recall at the expense of precision.

Table~\ref{tab:high-agreement} reports results on the high-agreement subset. 
High coverage and strong performance in this setting show that the first stage can reliably resolve many claims. 
However, when evidence is sparse or contradictory, the second-stage debate provides the additional reasoning needed, underscoring its critical role in the framework.

% This is also infer that the lower performance on SciFact with our two-stage method. The high coverage and high F1 score on high-agreement subset, which means in open domain articles are consistent. But when fallback to second-stage debate, it would be...

\begin{table}[h]
\centering
\small % or \footnotesize, \scriptsize
\begin{tabular}{|l|c|c|c|}
\hline
 & \textbf{SciFact} & \textbf{TREC-Health} & \textbf{HealthFC} \\
\hline
Coverage & 64.9\% & 50.1\% & 58.1\% \\
\hline
F1 Score & 92.0 & 88.6 & 84.0 \\
\hline
\end{tabular}
\caption{Results on the high-agreement subset. 
\textit{Coverage (\%)} denotes the proportion of claims settled without debate in the first stage, while \textit{F1 Score} reports the score for those claims.}
\label{tab:high-agreement}
\end{table}

% \subsection{Analysis} 

% We also measure the average number of debate rounds: $2.2$ in SciFact, $2.1$ in TREC-Health, and $2.4$ in HealthFC. These results indicate that, in most cases, only a single exchange of statements is sufficient for the judge to confidently determine the more persuasive side.

\section{Conclusion}
We proposed a two-stage framework for health misinformation detection that combines agreement score prediction with multi-agent debate. The first stage leverages weighted agreement scoring to resolve claims directly, while the second stage provides explainable reasoning through debate.  

Experiments on three health datasets demonstrate consistent improvements over strong baselines, including gains of $+0.8$ F1 on TREC-Health and $+1.4$ F1 on HealthFC, with a better balance between precision and recall.  
These results underscore the value of integrating evidence consistency with structured debate, advancing reliable and explainable health misinformation detection.  

\section*{Limitations}

While our two-stage framework achieves strong performance, it also entails certain limitations.  
First, as the approach relies on LLMs, the debate judge may still be affected by model biases or occasional hallucinations.  
Second, the multi-agent design requires multiple API calls, introducing extra computational cost; however, this cost is modest compared to the performance gains.  
Finally, our current evaluation is limited to binary-labeled datasets. Extending the framework to more nuanced settings, such as incorporating a \emph{Not Enough Information} class, represents a promising direction for future work. 

% \section*{Acknowledgments}

% \bibliography{custom}
% \bibliographystyle{acl_natbib}
\bibliography{references}

\appendix

\end{document}